\documentclass{article}

\usepackage{microtype}
\usepackage{graphicx}
\usepackage{subcaption}
\usepackage{booktabs} 
\usepackage{subfiles}
\usepackage{multirow}
\usepackage{wrapfig}
\usepackage{hyperref}
\usepackage{algorithm}
\usepackage{algorithmic}

\usepackage{bm}
\usepackage{subfiles}
\usepackage{wrapfig}
\usepackage{epsfig}
\usepackage{caption}

\graphicspath{{./images/}}
\DeclareGraphicsExtensions{.pdf,.jpeg,.jpg,.eps,.png}

\newcommand{\INST}[1]{}

\newcommand{\KL}[1]{\textbf{KL}}

\usepackage{amsmath, amsthm, amssymb, xspace}

\usepackage{color}
\definecolor{blue}{rgb}{0,0,.7}
\definecolor{red}{rgb}{.7,0,0}
\definecolor{orange}{rgb}{1,.6,0}
\definecolor{purple}{rgb}{.4,0,.5}
\definecolor{brown}{rgb}{.4,.2,.1}
\definecolor{green}{rgb}{0,.5,0}

\newcommand{\brck}[1]{\left(#1\right)}
\newcommand{\brcksq}[1]{\left[#1\right]}
\newcommand{\brckcur}[1]{\left\{#1\right\}}

\newcommand{\fr}[2]{\frac{#1}{#2}}

\newcommand{\be}{\begin{equation}}
\newcommand{\ee}{\end{equation}}
\newcommand{\bali}{\begin{eqnarray*}}
\newcommand{\eali}{\end{eqnarray*}}
\newcommand{\eq}[1]{\begin{align}#1\end{align}}

\newcommand{\iitem}[1]{\begin{itemize}#1\end{itemize}}

\newcommand{\calO}{\mathcal{O}}

\newcommand{\mathR}{\mathbb{R}}

\newcommand{\mathE}{\mathbb{E}}

\newcommand{\algMMT}{\texttt{MRTL}}
\newcommand{\algSGD}{\texttt{Opt}}
\newcommand{\algFinegrain}{\texttt{Finegrain}}
\newcommand{\algTensorFactorize}{\texttt{CP\_ALS}}

\newcommand{\full}{full rank }
\newcommand{\low}{low rank }

 \usepackage{mathtools}
 \usepackage{color}

\newcommand{\eat}[1]{}

\newtheorem{theorem}{Theorem}[section]

\newtheorem{definition}[theorem]{Definition}
\newtheorem{lemma}[theorem]{Lemma}

\newcommand{\E}{{\mathbb E}}

\newcommand{\T}[1]{{\mathcal{#1}}} 
\newcommand{\V}[1]{{\mathbf{#1}}} 

\makeatletter
\newcommand*{\rom}[1]{\expandafter\@slowromancap\romannumeral #1@}
\makeatother

\usepackage[accepted]{icml2020}

\icmltitlerunning{Multiresolution Tensor Learning for Efficient and Interpretable
 Spatial Analysis}

\begin{document}

\twocolumn[
\icmltitle{Multiresolution Tensor Learning for Efficient \\
and Interpretable Spatial Analysis}



\icmlsetsymbol{equal}{*}

\begin{icmlauthorlist}
\icmlauthor{Jung Yeon Park}{neu}
\icmlauthor{Kenneth Theo Carr}{neu}
\icmlauthor{Stephan Zheng}{sf}
\icmlauthor{Yisong Yue}{cal}
\icmlauthor{Rose Yu}{neu,ucsd}
\end{icmlauthorlist}

\icmlaffiliation{neu}{Khoury College of Computer Sciences, Northeastern University, Boston, MA, USA.}
\icmlaffiliation{sf}{Salesforce AI Research, Palo Alto, CA, USA. Work done while at California Institute of Technology.}
\icmlaffiliation{cal}{Department of Computing and Mathematical Sciences, California Institute of Technology, Pasadena, CA, USA.}
\icmlaffiliation{ucsd}{Computer Science and Engineering, University of California, San Diego, San Diego, CA, USA.}

\icmlcorrespondingauthor{Jung Yeon Park, Rose Yu}{\texttt{park.jungy@northeastern.edu}, \texttt{roseyu@eng.ucsd.edu} }

\icmlkeywords{Tensor models, multiresolution, multigrid, latent factors, spatial analysis}

\vskip 0.3in
]



\printAffiliationsAndNotice{}  

\begin{abstract}
Efficient and interpretable spatial analysis is crucial in many fields such as geology, sports, and climate science. Tensor latent factor models can describe higher-order correlations for spatial data. However, they are computationally expensive to train and are sensitive to initialization, leading to spatially incoherent, uninterpretable results. We develop a novel Multiresolution Tensor Learning (\algMMT{}) algorithm for efficiently learning interpretable spatial patterns. \algMMT{} initializes the latent factors from an approximate full-rank tensor model for improved interpretability and progressively learns from a coarse resolution to the fine resolution to reduce computation. We also prove the theoretical convergence and computational complexity of \algMMT{}. When applied to two real-world datasets, \algMMT{} demonstrates $4 \sim 5$x speedup compared to a fixed resolution approach while yielding accurate and interpretable latent factors.
\end{abstract}

\section{Introduction}
\label{sec:intro}
Analyzing large-scale spatial data plays a critical role in sports, geology, and climate science. In spatial statistics, kriging or Gaussian processes are popular tools for spatial analysis  \cite{cressie1992statistics}. Others have proposed various Bayesian methods such as Cox processes \cite{Miller2014,dieng2017variational} to model spatial data. However, while mathematically appealing, these methods often have difficulties scaling to high-resolution data. 

\begin{wrapfigure}{tr}{0.5\linewidth}
\centering
    \begin{subfigure}[t]{0.24\columnwidth}
        \includegraphics[width=\textwidth]{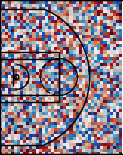}
    \end{subfigure}
    \begin{subfigure}[t]{0.24\columnwidth}
        \includegraphics[width=\textwidth]{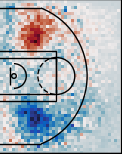}
    \end{subfigure}
\vspace{-0.1in}
\caption{
Latent factors: random (left) vs. good (right) initialization. Latent factors vary in interpretability depending on initialization.}
\label{fig:intro_rand_lfs}
\end{wrapfigure}

We are interested in learning high-dimensional tensor latent factor models, which have shown to be a scalable alternative for spatial analysis \cite{yu2018tensor,litvinenko2019tucker}.  High resolution spatial data often contain higher-order correlations between features and locations, and tensors can naturally encode such multi-way correlations.
For example, in competitive basketball play, we can predict how each player's decision to shoot is jointly influenced by their shooting style, his or her court position, and the position of the defenders by simultaneously encoding these features as a tensor. Using such representations,  learning tensor latent factors can directly extract higher-order correlations.

A challenge in such models is high computational cost. High-resolution spatial data is often discretized, leading to large high-dimensional tensors whose training scales exponentially with the number of parameters. Low-rank tensor learning \cite{yu2018tensor, kossaifi2019tensorly} reduces the dimensionality by assuming low-rank structures in the data and uses tensor decomposition to discover latent  semantics; for an overview of tensor learning, see review papers \cite{Kolda2009, sidiropoulos2017tensor}. However, many tensor learning methods have been shown to be sensitive to noise \cite{cheng2016scalable} and initialization \cite{anandkumar2014guaranteed}. Other numerical techniques, including random sketching \cite{wang2015fast,haupt2017near} and parallelization, \cite{austin2016parallel,li2017model} can speed up training, but they often fail to utilize the unique properties of spatial data such as spatial auto-correlations. 


Using latent factor models also gives rise to another issue: interpretability. It is well known that a latent factor model is generally not identifiable \cite{allman2009identifiability}, leading to uninterpretable factors that do not offer insights to domain experts. In general, the definition of interpretability is highly application dependent \cite{doshi2017towards}.  For spatial analysis, one of the unique properties of spatial patterns is \textit{spatial auto-correlation}: close objects have similar values \cite{moran1950notes}, which we use as a criterion for interpretability. As latent factor models are sensitive to initialization, previous research \cite{Miller2014, Yue2014} has shown that randomly initialized latent factor models can lead to spatial patterns that violate spatial auto-correlation and hence are not interpretable (see Fig. \ref{fig:intro_rand_lfs}).

In this paper, we propose a Multiresolution Tensor Learning algorithm, \algMMT{}, to efficiently learn accurate and interpretable patterns in spatial data. \algMMT{} is based on two key insights. First, to obtain good initialization, we train a full-rank tensor model approximately at a low resolution and use tensor decomposition to produce latent factors. Second, we exploit spatial auto-correlation to learn models at multiple resolutions: we train starting from a coarse resolution and iteratively finegrain to the next resolution.  
We provide theoretical analysis and prove the convergence properties and computational complexity of \algMMT{}. 
We demonstrate on two real-world datasets that this approach is significantly faster than fixed resolution methods.
We develop several finegraining criteria  to determine when to finegrain. We also consider different interpolation schemes and discuss how to finegrain in different applications. The code for our implementation is available \footnote{ \url{https://github.com/Rose-STL-Lab/mrtl}}.

In summary, we:
\iitem{
\itemsep0em 
\item propose a Multiresolution  Tensor Learning (\algMMT{}) optimization algorithm  for large-scale spatial analysis.
\item prove the rate of convergence for \algMMT{} which depends on the spectral norm of the interpolation operator. We also show the exponential computational speedup for \algMMT{} compared with fixed resolution. 
\item develop different criteria to determine when to transition to a finer resolution and discuss different finegraining methods.
\item evaluate on two real-world datasets and show \algMMT{} learns faster than fixed-resolution learning and can produce interpretable latent factors.
}

\section{Related Work.}
\paragraph{Spatial Analysis}
Discovering spatial patterns has significant implications in scientific fields such as human behavior modeling, neural science, and climate science. Early work in spatial statistics has contributed greatly to spatial analysis through the work in Moran's I \cite{moran1950notes} and Getis-Ord general G \cite{getis1992analysis} for measuring spatial auto-correlation. Geographically weighted regression \cite{brunsdon1998geographically} accounts for the spatial heterogeneity with a local version of spatial regression but fails to capture higher order correlation. Kriging or Gaussian processes are popular tools for spatial analysis but they often require carefully designed variograms (also known as kernels) \cite{cressie1992statistics}. Other Bayesian hierarchical models favor spatial point processes to model spatial data \cite{diggle2013spatial,Miller2014,dieng2017variational}. These frameworks are conceptually elegant but often computationally intractable. 
%
\paragraph{Tensor Learning}
Latent factor models utilize correlations in the data to reduce the dimensionality of the problem, and have been used extensively in multi-task learning  \cite{romera2013multilinear} and recommendation systems \cite{lee2001algorithms}. Tensor learning \cite{zhou2013tensor, bahadori2014fast, haupt2017near} uses tensor latent factor models to learn higher-order correlations in the data in a supervised fashion. In particular, tensor latent factor models aim to learn the higher-order correlations in spatial data by assuming low-dimensional representations among features and locations. 
However, high-order tensor models are non-convex by nature, suffer from the curse of dimensionality, and are notoriously hard to train \cite{Kolda2009,sidiropoulos2017tensor}.
There are many efforts to scale up tensor computation, e.g., parallelization \citep{austin2016parallel} and sketching \citep{wang2015fast, haupt2017near, li2018sketching}. In this work, we propose an optimization algorithm to learn tensor models at multiple resolutions that is not only fast but can also generate interpretable factors. We focus on tensor latent factor models for their wide applicability to spatial analysis and interpretability. While deep neural networks models can be more accurate, they are computationally more expensive and are difficult to interpret.
\paragraph{Multiresolution Methods}
Multiresolution methods have been applied successfully in machine learning, both in latent factor modeling \cite{kondor2014multiresolution,ozdemir2017multiscale} and deep learning \cite{reed2017parallel,serban2017multiresolution}. For example, multiresolution matrix factorization \citep{kondor2014multiresolution, ding2017multiresolution} and its higher order extensions \citep{schifanella2014multiresolution, ozdemir2017multiscale, han2018multiresolution} apply multi-level orthogonal operators to uncover the multiscale structure in a single matrix. In contrast, our method aims to speed up learning by exploiting the relationship among multiple tensors of different resolutions.  Our approach resembles the multigrid method in numerical analysis for solving partial differential equations \cite{trottenberg2000multigrid, hiptmair1998multigrid}, where the idea is to accelerate iterative algorithms by solving a coarse problem first and then gradually finegraining the solution.

\section{Tensor Models for Spatial Data}
\label{sec:model}

We consider  tensor learning in the supervised setting. We describe both models for the full-rank case and the low-rank case. An order-3 tensor is used for ease of illustration but our model covers higher order cases.

\subsection{Full Rank Tensor Models}
%
Given input data consisting of both non-spatial and spatial features, we can discretize the spatial features at $r = 1, \dots, R$ resolutions, with corresponding dimensions as $D_1, \dots, D_R$.
%
Tensor learning parameterizes the model with a weight tensor $\T{W}^{(r)} \in \mathR^{I \times F \times D_r}$  over all features, where $I$ is number of outputs and $F$ is number of non-spatial features. The input data is of the form $\T{X}^{(r)} \in \mathbb{R}^{I\times F \times D_r}$. Note that both the input features and the learning model are resolution dependent. $\T{Y}_i \in \mathR, i=1,\dots,I$ is the label for output $i$. 

At resolution $r$, the full rank tensor learning model can be written as
\begin{eqnarray}
\label{eq:class-full}
\T{Y}_{i} = a\left(\sum_{f=1}^F \sum_{d=1}^{D_r} \T{W}_{i,f,d}^{(r)} \T{X}^{(r)}_{i,f, d} + b_i\right) \text{,}
\end{eqnarray}
where $a$ is the activation function and $b_i$ is the bias for output $i$. The weight tensor $\T{W}$ is contracted with $\T{X}$ along the non-spatial mode $f$ and the spatial mode $d$.  In general, Eqn. \eqref{eq:class-full} can be extended to multiple spatial features and spatial modes, each of which can have its own set of resolution-dependent dimensions. We use a sigmoid activation function for the classification task and the identity activation function for regression. 


\subsection{Low Rank Tensor Model}
Low rank tensor models assume a low-dimensional latent structure in $\T{W}$ which can characterize distinct patterns in the data and also alleviate model overfitting. To transform the learned tensor model to a low-rank one, we use CANDECOMP/PARAFAC (CP) decomposition \cite{hitchcock1927expression} on $\T{W}$, which assumes that $\T{W}$ can be represented as the sum of rank-1 tensors. Our method can easily be extended for other decompositions as well. 

Let $K$ be the CP rank of the tensor. In practice, $K$ cannot be found analytically and is often chosen to sufficiently approximate the dataset. The weight tensor $\T{W}^{(r)}$ is factorized into multiple factor matrices as
\[\T{W}_{i,f,d}^{(r)} = \sum_{k=1}^K A_{i,k} B_{f,k} C^{(r)}_{d,k} \]
 The tensor latent factor model is
\begin{eqnarray}
\label{eq:class-low}
\T{Y}_{i} = a \left( \sum_{f=1}^F \sum_{d=1}^{D_r} \sum_{k=1}^{K} A_{i,k} B_{f,k} C^{(r)}_{d,k}\T{X}^{(r)}_{i,f, d} + b_i\right) \text{,}
\end{eqnarray}
where the columns of $A, B, C^r$ are latent factors for each mode of $\T{W}$ and $C^{(r)}$ is resolution dependent.

CP decomposition reduces dimensionality by assuming that $A, B, C^r$ are uncorrelated, i.e. the features are uncorrelated. This is a reasonable assumption depending on how the features are chosen and leads to enhanced spatial interpretability as the learned spatial latent factors can show common patterns regardless of other features.

\subsection{Spatial Regularization}
Interpretability is in general hard to define or quantify \cite{doshi2017towards, ribeiro2016should, lipton2018mythos, molnar2019}. In the context of spatial analysis, we deem a latent factor as interpretable if it produces a spatially coherent pattern exhibiting spatial auto-correlation. To this end, we utilize a spatial regularization kernel \cite{lotte2010regularizing, Miller2014, Yue2014} and extend this to the tensor case. 

Let $d = 1,\dots,D_r$ index all locations of the spatial dimension for resolution $r$. The spatial regularization term is:
\begin{equation}
    \label{eq:spatial-reg}
    R_s = \sum_{d=1}^{D_r} \sum_{d'=1}^{D_r} K_{d, d'} \| \T{W}_{:, :, d} - \T{W}_{:, :, d'}\|_F^2 \text{ ,}
\end{equation}
where $\| \cdot \|_F$ denotes the Frobenius norm and $K_{d, d'}$ is the kernel that controls the degree of similarity between locations. We use a simple RBF kernel with hyperparameter $\sigma$.
\begin{equation}
    \label{eq:spatial-rbf}
    K_{d, d'} = e^{(-\|l_d - l_{d'}\|^2 / \sigma)} \text{ ,}
\end{equation}
where $l_d$ denotes the location of index $d$. The distances are normalized across resolutions such that the maximum distance between two locations is $1$. The kernels can be precomputed for each resolution. If there are multiple spatial modes, we apply spatial regularization across all different modes. We additionally use $L_2$ regularization to encourage smaller weights. The optimization objective function  is 
\begin{equation}
   f(\T{W}) = L(\T{W};\T{X},\T{Y}) + \lambda_R R(\T{W}) \text{ ,}
\end{equation}
where $L$ is a task-dependent supervised learning loss, $R(\T{W})$ is the sum of spatial and $L_2$ regularization, and $\lambda_R$ is the regularization coefficient. 
\section{Multiresolution Tensor Learning}
\label{sec:mrtl}

We now describe our algorithm \algMMT{}, which addresses both the computation and interpretability issues. Two key concepts of \algMMT{} are learning good initializations and utilizing multiple resolutions.

\subsection{Initialization}
In general, due to their nonconvex nature, tensor latent factor models are sensitive to initialization and can lead to uninterpretable latent factors \cite{Miller2014, Yue2014}. We use full-rank initialization in order to learn latent factors that correspond to known spatial patterns.

We first train an approximate full-rank version of the tensor model at a low resolution in Eqn. \eqref{eq:class-full}. The weight tensor is then decomposed into latent factors and these values are used to initialize the low-rank model. The low-rank model in Eqn. \eqref{eq:class-low} is then trained to the final desired accuracy. As we use approximately optimal solutions of the full-rank model as initializations for the low-rank model, our algorithm produces interpretable latent factors in a variety of different scenarios and datasets. 

Full-rank initialization requires more computation than other simpler initialization methods. However, as the full-rank model is trained only for a small number of epochs, the increase in computation time is not substantial. We also train the full-rank model only at lower resolutions, for further reduction. 

Previous research \cite{Yue2014} showed that spatial regularization alone is not enough to learn spatially coherent factors, whereas full-rank initialization, though computationally costly, is able to fix this issue. We confirm the same holds true in our experiments (see Section \ref{sub:random_init}). Thus, full-rank initialization is critical for spatial interpretability.

\subsection{Multiresolution}
Learning a high-dimensional tensor model is generally computationally expensive and memory inefficient. We utilize multiple resolutions for this issue. We outline the procedure of \algMMT{} in Alg. \ref{alg:mrtl}, where we omit the bias term in the description for clarity. 

\begin{algorithm}[t]
\caption{Multiresolution Tensor Learning: \algMMT{}}
	\label{alg:mrtl}
	\begin{small}
		\begin{algorithmic}[1]
\STATE Input: initialization $\T{W}_{0}$, data $\T{X},\T{Y}$.
\STATE Output: latent factors $\T{F}^{(r)}$
\STATE \textit{\# full rank tensor model}
\FOR{each resolution $r \in \brckcur{1, \ldots, r_0}$}
    \STATE Initialize $t \gets 0$
    \STATE Get a mini-batch $\T{B}$ from training set
    \WHILE{stopping criterion not true}
        \STATE $t \gets t + 1$
	    \STATE $\T{W}^{(r)}_{t+1} \gets$ \algSGD{}$\brck{\T{W}^{(r)}_{t} \mid \T{B}}$
    \ENDWHILE
    \STATE  $\T{W}^{(r+1)}$ = \algFinegrain{}$\brck{\T{W}^{(r)}}$
\ENDFOR
\STATE \textit{\# tensor decomposition}
\STATE $\T{F}^{(r_0)} \gets$ \algTensorFactorize{}$\brck{\T{W}^{(r_0)}}$
\STATE \textit{\# low rank tensor model}
\FOR{each resolution $r \in \brckcur{r_0, \ldots, R}$}
    \STATE Initialize $t \gets 0$
    \STATE Get a mini-batch $\T{B}$ from training set
    \WHILE{stopping criterion not true}
        \STATE $t \gets t + 1$
	    \STATE $\T{F}^{(r)}_{t+1} \gets$ \algSGD{}$\brck{\T{F}^{(r)}_{t} \mid \T{B}}$
    \ENDWHILE
    \FOR{each spatial factor $n\in \{1,\cdots, N\}$}
        \STATE $\T{F}^{(r+1),n}$ = \algFinegrain{}$\brck{\T{F}^{(r),n}}$
    \ENDFOR
\ENDFOR
\end{algorithmic}
\end{small}
\end{algorithm}

We represent the resolution $r$ with superscripts and the iterate at step $t$ with subscripts, i.e. $\T{W}^{(r)}_{t}$ is $\T{W}$ at resolution $r$ at step $t$. $\T{W}_0$ is the initial weight tensor at the lowest resolution. $\T{F}^{(r)}=(A,B,C^{(r)})$ denotes all factor matrices at resolution $r$ and we use $n$ to index the factor $\T{F}^{(r), n}$.

For efficiency, we train both the full rank and low rank models at multiple resolutions, starting from a coarse spatial resolution and progressively increase the resolution. At each resolution $r$, we learn $\T{W}^{(r)}$ using the stochastic optimization algorithm of choice $\algSGD{}$ (we used Adam \citep{kingma2014adam} in our experiments). When the stopping criterion is met, we transform $\T{W}^{(r)}$ to $\T{W}^{(r+1)}$ in a process we call finegraining (\algFinegrain{}). Due to spatial auto-correlation, the trained parameters at a lower resolution will serve as a good initialization for higher resolutions. For both models, we only finegrain the factors that corresponds to resolution dependent mode, which is the spatial mode in the context of spatial analysis. Finegraining can be done for other non-spatial modes for more computational speedup as long as there exists a multiresolution structure (e.g. video or time series data).

Once the full rank resolution has been trained up to resolution $r_0$ (which can be chosen to fit GPU memory or time constraints), we decompose $\T{W}^{(r)}$ using \algTensorFactorize{}, the standard alternating least squares (ALS) algorithm \cite{Kolda2009} for CP decomposition. Then the low-rank model is trained at resolutions $r_0,\dots,R$ to final desired accuracy, finegraining to move to the next resolution.

\paragraph{When to finegrain}
There is a tradeoff between training times at different resolutions. While training for longer at lower resolutions significantly decreases computation, we do not want to overfit to the coarse, lower resolution data. On the other hand, training at higher resolutions can yield more accurate solutions using more detailed information. We investigate four different criteria to balance this tradeoff: 1) validation loss, 2) gradient norm, 3) gradient variance, and 4) gradient entropy.

Increase in validation loss \cite{prechelt1998automatic, yao2007early} is a commonly used heuristic for early stopping. Another approach is to analyze the gradient distributions during training. For a convex function, stochastic gradient descent will converge into a noise ball near the optimal solution as the gradients approach zero. However, lower resolutions may be too coarse to learn more finegrained curvatures and the gradients will increasingly disagree near the optimal solution. We quantify the disagreement in the gradients with metrics such as norm, variance, and entropy. We use intuition from convergence analysis for gradient norm and variance \cite{bottou2018optimization}, and information theory for gradient entropy \cite{srinivas2012information}. 

Let $\T{W}_t$ and $\xi_t$ represent the weight tensor and the random variable for sampling of minibatches at step $t$, respectively. Let $f(\T{W}_t; \xi_t) := f_t$ be the validation loss and $g(\T{W}_t; \xi_t) := g_t$ be the stochastic gradients at step $t$. The finegraining criteria are:
\begin{itemize}
    \item Validation Loss: \small{$\mathE[f_{t+1}] - \mathE[f_t] > 0$}
    \item Gradient Norm: \small{$\mathE[\|g_{t+1}\|^2] - \mathE[\|g_t\|^2] > 0$}
    \item Gradient Variance: \small{$V(\mathE[g_{t+1}]) - V(\mathE[g_t]) > 0$}
    \item Gradient Entropy: \small{$S(\mathE[g_{t+1}]) - S(\mathE[g_{t}]) > 0$} \text{,}
\end{itemize}
where $S(p) = \sum_{i} -p_i \ln(p_i)$. One can also use thresholds, e.g. $|f_{t+1} - f_{t}| < \tau$, but as these are dependent on the dataset, we use $\tau=0$ in our experiments. One can also incorporate patience, i.e. setting the maximum number of epochs where the stopping conditions was reached.

\paragraph{How to finegrain}
We discuss different interpolation schemes for different types of features. Categorical/multinomial variables, such as a player's position on the court, are one-hot encoded or multi-hot encoded onto a discretized grid. Note that as we use higher resolutions, the sum of the input values are still equal across resolutions, $\sum_{d} \T{X}^{(r)}_{:,:,d} = \sum_{d} \T{X}^{(r+1)}_{:,:,d}$. As the sum of the features remains the same across resolutions and our tensor models are multilinear, nearest neighbor interpolation should be used in order to produce the same outputs. 
\begin{align*}
    \sum_{d=1}^{D_{r}} \T{W}_{:,:,d}^{(r)} \T{X}^{(r)}_{:,:, d} = \sum_{d=1}^{D_{r+1}} \T{W}_{:,:,d}^{(r+1)} \T{X}^{(r+1)}_{:,:, d}
\end{align*}
as $\T{X}^{(r)}_{i,f, d} = 0$ for cells that do not contain the value. This scheme yields the same outputs and thus the same loss values across resolutions.

Continuous variables that represent averages over locations, such as sea surface salinity, often have similar values at each finegrained cell at higher resolutions (as the values at coarse resolutions are subsampled or averaged from values at the higher resolution).
Then $\sum_{d}^{D_{r+1}} \T{X}^{(r+1)}_{:,:,d} \approx 2^2 \sum_{d}^{D_r} \T{X}^{(r)}_{:,:,d}$, where the approximation comes from the type of downsampling used.
\begin{align*}
    \sum_{d=1}^{D_{r}} \T{W}_{:,:,d}^{(r)} \T{X}^{(r)}_{:,:, d} \approx 2^2 \sum_{d=1}^{D_{r+1}} \T{W}_{:,:,d}^{(r+1)} \T{X}^{(r+1)}_{:,:, d}
\end{align*}
using a linear interpolation scheme. The weights are divided by the scale factor of $\frac{D_{r+1}}{D_{r}}$ to keep the outputs approximately equal. We use bilinear interpolation, though any other linear interpolation can be used.

\section{Theoretical Analysis.} 
\subsection{Convergence}
We  prove the convergence rate for \algMMT{} with a single spatial mode and one-dimensional output, where the weight tensor reduces to a weight vector $\V{w}$.  We defer all proofs to  Appendix \ref{supp:theory}. For the loss function $f$ and a stochastic sampling variable $\xi$,
%
the optimization problem is:
\begin{equation}
    \V{w}_\star = \text{argmin}\ \E[f(\V{w};\xi)]
\end{equation}
We consider a fixed-resolution model that follows Alg. \ref{alg:mrtl} with $r=\{R\}$, i.e. only the final resolution is used. For a fixed-resolution miniSGD algorithm, under common assumptions in convergence analysis:
\begin{itemize}
    \item  $f$ is $\mu$- strongly convex, $L$-smooth 
    \item (unbiased) gradient $\E[g(\V{w}_t; \V{\xi}_t)] =\triangledown f(\V{w}_t)$ given $\xi_{<t}$
    \item (variance) for all the $\V{w}$, $\E[\|g(\V{w}; \xi)\|^2_2] \leq \sigma_g^2 + c_g \|\triangledown f(\V{w})\|_2^2 $
\end{itemize}
\begin{theorem}\cite{bottou2018optimization}
If the step size $\eta_t \equiv \eta \leq \frac{1}{Lc_g}$, then a fixed resolution solution satisfies
\begin{align*}
    \E[\|\V{w}_{t+1} - \V{w}_{\star}\|^2_2] \leq & \gamma^t(\E[\|\V{w}_{0} - \V{w}_{\star}\|^2_2) - \beta] 
     + \beta \text{,}
\end{align*}
where $\gamma = 1 - 2 \eta \mu$, $\beta=\frac{\eta \sigma_g^2}{2\mu}$, and $w_\star$ is the optimal solution.
\end{theorem}
which gives $O(1/t)+O(\eta)$ convergence.


At resolution $r$, we define the number of total iterations as $t_r$, and the weights as $\V{w}^{(r)}$. We let $D_r$ denote the number of dimensions at $r$ and we assume a dyadic scaling between resolutions such that $D_{r+1}=2D_r$. We define finegraining using an interpolation operator $P$ such that $\V{w}^{(r+1)}_{0} = P \V{w}^{(r)}_{t_r}$ as in \cite{bramble2019multigrid}. For the simple case of a 1D spatial grid where $\V{w}_t^{(r)}$ has spatial dimension $D_r$, P would be of a Toeplitz matrix of dimension $2D_r \times D_r$. For example, for linear interpolation of $D_r=2$,
\begin{align*}
    P \V{w}^{(r)} = \frac{1}{2}\begin{bmatrix}
    1 & 0 \\
    2 & 0 \\
    1 & 1 \\
    0 & 2 \\
\end{bmatrix} \begin{bmatrix} \V{w}_1^{(r)} \\ \V{w}_2^{(r)} \end{bmatrix} = \begin{bmatrix} \V{w}_1^{(r+1)}/2 \\ \V{w}_1^{(r+1)} \\ \V{w}_1^{(r+1)}/2 + \V{w}_2^{(r+1)}/2 \\ \V{w}_2^{(r+1)} \end{bmatrix} \text{.}
\end{align*}
Any interpolation scheme can be expressed in this form.

The convergence of multiresolution learning algorithm depends on the following  property of spatial data:
\begin{definition}[Spatial Smoothness]
The difference between the optimal solutions of consecutive resolutions is upper bounded by $\epsilon$ 
\[\| \V{w}^{(r+1)}_\star -  P \V{w}^{(r)}_\star \| \leq \epsilon \text{,}\]
with $P$ being the interpolation operator.
\end{definition}
The following theorem proves the convergence rate of \algMMT{}, with a constant that depends on the operator norm of the interpolation operator $P$. 
\begin{theorem} If the step size $\eta_t \equiv \eta \leq \frac{1}{Lc_g}$, then the solution of \algMMT{} satisfies
\begin{align*}
    &\E[\|\V{w}^{(r)}_{t} - \V{w}_{\star}\|^2_2] \leq \gamma^{t} \|P\|^{2r}_{op}~\E[\|\V{w}_{0} - \V{w}_\star\|^2_2  + O(\eta \|P\|_{op}) \text{,}
\end{align*}
where $\gamma = 1 - 2 \eta \mu$, $\beta=\frac{\eta \sigma_g^2}{2\mu}$, and $\|P\|_{op}$ is the operator norm of the interpolation operator $P$.
\end{theorem}

\subsection{Computational Complexity}
To analyze computational complexity, we resort to fixed point convergence \citep{hale2008fixed} and the multigrid method \citep{stuben2001review}. Intuitively, as most of the training iterations are spent on coarser resolutions with fewer number of parameters, multiresolution learning is more efficient than fixed-resolution training.

Assuming that $\nabla f$ is Lipschitz continuous, we can view gradient-based optimization as a fixed-point iteration operator $F$ with a contraction constant of $\gamma \in (0,1)$ (note that \emph{stochastic} gradient descent converges to a noise ball instead of a fixed point):
\begin{eqnarray*}
	\V{w} \leftarrow F(\V{w}), \hspace{10pt} F:=I-\eta \nabla f,  \\
	\| F(\V{w} ) - F(\V{w}') \|\leq \gamma \| \V{w} - \V{w}'\|
\end{eqnarray*}
Let $\V{w}^{(r)}_\star$ be the optimal estimator at resolution $r$ and  $\V{w}^{(r)}$ be a solution satisfying $\| \V{w}^{(r)}_\star -\V{w}^{(r)}\| \leq \epsilon/2$. The algorithm terminates when the estimation error reaches $\fr{C_0 R}{(1-\gamma)^2}$. The following lemma describes the computational cost of the \emph{fixed-resolution} algorithm.
\begin{lemma}
Given a fixed point iteration operator $F$ with contraction constant of $\gamma \in (0,1)$, the computational complexity of fixed-resolution training for tensor model of order $p$ and rank $K$ is
	\eq{
		\mathcal{C} = \calO\brck{
			\fr{1}{|\log \gamma|} \cdot \log \left(\fr{1 }{(1-\gamma)\epsilon}\right)
			\cdot\fr{Kp}{(1-\gamma)^2\epsilon}} \text{,}
	} \label{lemma:fixed}
where $\epsilon$ is the terminal estimation error.
\end{lemma}
The next Theorem \ref{thm:mmt} characterizes the computational speed-up gained by \algMMT{} compared to fixed-resolution learning, with respect to the contraction factor $\gamma$ and the terminal estimation error $\epsilon$.
\begin{theorem}
\label{thm:mmt}
If the fixed point iteration operator (gradient descent) has a contraction factor of $\gamma$, multiresolution learning with the termination criteria of $\fr{C_0 r}{(1-\gamma)^2}$ at resolution $r$  is faster than fixed-resolution learning by a factor of  $ \log\fr{1}{(1-\gamma) \epsilon}$, with the terminal estimation error $\epsilon$.
\end{theorem}
Note that the speed-up using multiresolution learning uses a global convergence criterion $\epsilon$ for each $r$.

\section{Experiments}
\label{sec:experiments}

We apply \algMMT{} to two real-world datasets: basketball tracking and climate data. More details about the datasets and pre-processing steps are provided in Appendix \ref{supp:experiments}. 

\subsection{Datasets}
\paragraph{Tensor classification: Basketball tracking}
We use a large NBA player tracking dataset from \citep{Yue2014,zheng2016generating} consisting of the coordinates of all players at 25 frames per second, for a total of approximately 6 million frames. The goal is to predict whether a given ball handler will shoot within the next second, given his position on the court and the relative positions of the defenders around him. In applying our method, we hope to obtain common shooting locations on the court and how a defender's relative position suppresses shot probability.
\begin{figure}[h]
  \centering
	\includegraphics[width=0.8\linewidth]{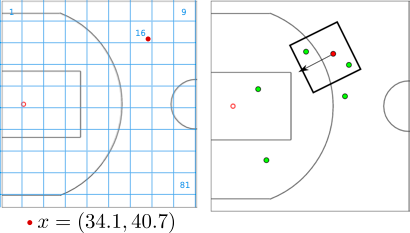}
	\captionof{figure}{Left: Discretizing a continuous-valued position of a player (red) via a spatial grid. Right: sample frame with a ballhandler (red) and defenders (green). Only defenders close to the ballhandler are used. }
\label{fig:basketball_data}

\end{figure}

The basketball data contains two spatial modes: the ball handler's position and the relative defender positions around the ball handler. We instantiate a tensor classification model in Eqn \eqref{eq:class-full} as follows:
\begin{eqnarray*}
\T{Y}_{i} =  \sum_{d^1=1}^{D^1_r} \sum_{d^2=1}^{D^2_r} \sigma(\T{W}_{i,d^1,d^2}^{(r)} \T{X}^{(r)}_{i, d^1,d^2} + b_i) \text{ ,}
\end{eqnarray*}
where $i \in \{1,\dots,I\}$ is the ballhandler ID, $d^1$ indexes the ballhandler's position on the discretized court of dimension $\{D_r^1\}$, and $d^2$ indexes the relative defender positions around the ballhandler in a discretized grid of dimension $\{D_r^2\}$. We assume that only defenders close to the ballhandler affect shooting probability and set $D_r^2 < D_r^1$ to reduce dimensionality. As shown in Fig. \ref{fig:basketball_data}, we orient the defender positions so that the direction from the ballhandler to the basket points up. $\T{Y}_{i} \in \{0, 1\}$ is the binary output equal to $1$ if player $i$ shoots within the next second and $\sigma$ is the sigmoid function.

We use nearest neighbor interpolation for finegraining and a weighted cross entropy loss (due to imbalanced classes):
\begin{equation}
\T{L}_n = -\beta \left[\T{Y}_n \cdot \log{\hat{\T{Y}}_n} + (1-\T{Y}_n)\cdot \log{(1 - \hat{\T{Y}}_n)}\right] \text{,}
\end{equation}
where $n$ denotes the sample index and $\beta$ is the weight of the positive samples and set equal to the ratio of the negative and positive counts of labels.

\paragraph{Tensor regression: Climate}

\begin{figure}[h]
  \centering
\includegraphics[width=0.95\linewidth]{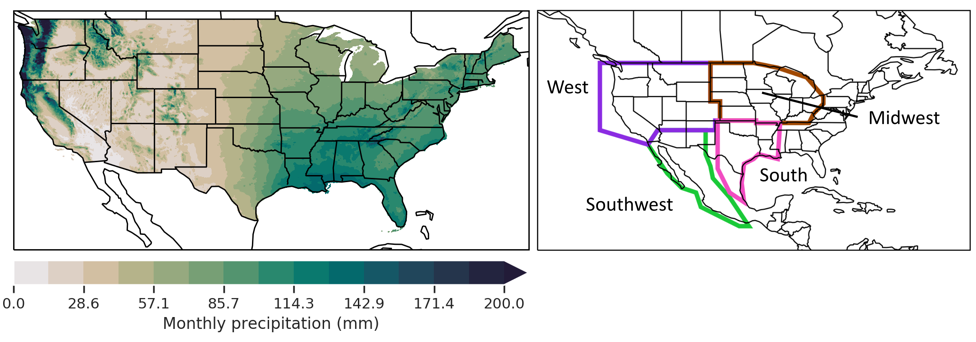}
	\captionof{figure}{Left: precipitation over continental U.S. Right: regions considered in particular.}
\label{fig:climate_data}
\end{figure}

Recent research \cite{li_2016_midwest, li_2016_sahel, zeng_2019_yangtze} shows that oceanic variables such as sea surface salinity (SSS) and sea surface temperature (SST) are significant predictors of the variability in rainfall in land-locked locations, such as the U.S. Midwest. We aim to predict the variability in average monthly precipitation in the U.S. Midwest using SSS and SST to identify meaningful latent factors underlying the large-scale processes linking the ocean and precipitation on land (Fig. \ref{fig:climate_data}). We use precipitation data from the PRISM group \cite{prism} and SSS/SST data from the EN4 reanalysis \cite{en4}.

Let $\T{X}$ be the historical oceanic data with spatial features SSS and SST across $D_r$ locations, using the previous $6$ months of data. As SSS and SST share the spatial mode (the same spatial locations), we set the $F_2=2$ to denote the index of these features. We also consider the lag as a non-spatial feature so that $F_1=6$. We instantiate the tensor regression model in Eqn \eqref{eq:class-full} as follows:
\begin{eqnarray*}
\T{Y} =  \sum_{f_1=1}^{F_1} \sum_{f_2=1}^{F_2} \sum_{d=1}^{D_r} \T{W}_{f_1, f_2, d}^{(r)} \T{X}^{(r)}_{f_1, f_2, d} + b
\end{eqnarray*}
The features and outputs (SSS, SST, and precipitation) are subject to long-term trends and a seasonal cycle. We use difference detrending for each timestep due to non-stationarity of the inputs, and remove seasonality in the data by standardizing each month of the year. The features are normalized using min-max normalization. We also normalize and deseasonalize the outputs, so that the model predicts standardized anomalies. We use mean square error (MSE) for the loss function and bilinear interpolation for finegraining.

\paragraph{Implementation Details}
For both datasets, we discretize the spatial features and use a 60-20-20 train-validation-test set split. We use Adam \citep{kingma2014adam} for optimization as it was empirically faster than SGD in our experiments. We use both $L_2$ and spatial regularization as described in Section \ref{sec:model}. We selected optimal hyperparameters for all models via random search. We use a stepwise learning rate decay with stepsize of $1$ with $\gamma=0.95$. We perform ten trials for all experiments. All other details are provided in Appendix \ref{supp:experiments}.

\subsection{Accuracy and Convergence}
We compare \algMMT{} against a fixed-resolution model on accuracy and computation time. We exclude the computation time for \algTensorFactorize{} as it was quick to compute for all experiments ($<5$ seconds for the basketball dataset). The results of all trials are listed in Table \ref{tab:results}. Some results are provided in Appendix \ref{supp:experiments}.

\begin{figure}[t]
    \centering
    \begin{subfigure}[t]{0.49\linewidth}
        \includegraphics[width=\textwidth]{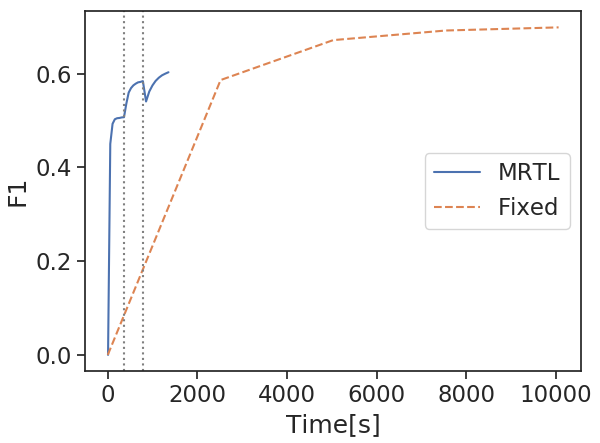}
    \end{subfigure}
    \begin{subfigure}[t]{0.49\linewidth}
        \includegraphics[width=\textwidth]{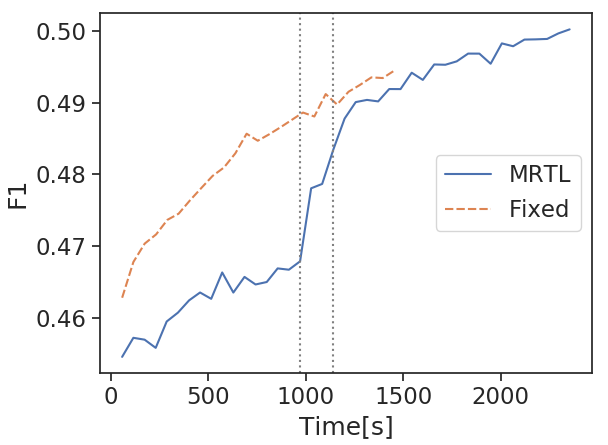}
    \end{subfigure}
\caption{Basketball: F1 scores of \algMMT{} vs. the fixed-resolution model for the \full (left) and \low model (right). The vertical lines indicate finegraining to the next resolution.}
\label{fig:bball_multi_fixed}
\end{figure}

\begin{figure}[t]
    \centering
    \begin{subfigure}[t]{0.49\columnwidth}
        \includegraphics[width=\textwidth]{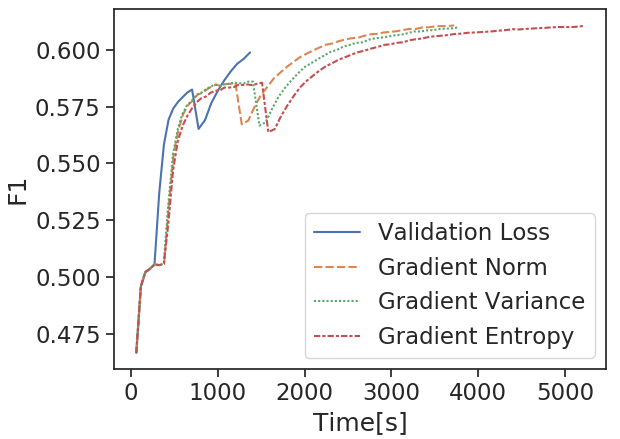}
    \end{subfigure}
    \begin{subfigure}[t]{0.49\columnwidth}
        \includegraphics[width=\textwidth]{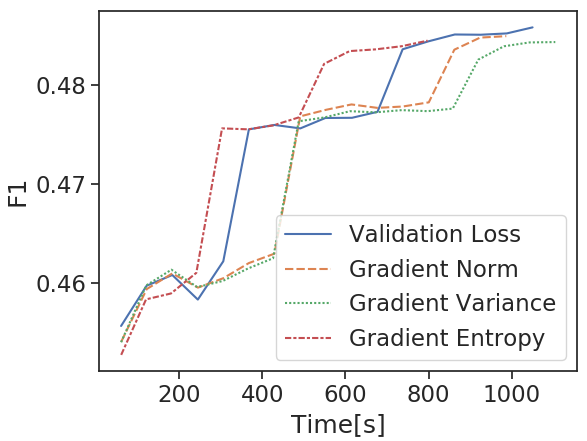}
\end{subfigure}
\caption{Basketball: F1 scores different finegraining criteria for the \full (left) and \low (right) model}
\label{fig:bball_stop_cond}
\end{figure}

\begin{table*}[t]
\footnotesize
\centering
\captionsetup{justification=centering}
\caption{Runtime and prediction performance comparison of a fixed-resolution model vs \algMMT{} for datasets}
\begin{tabular}{l|l|c|c|c|c|c|c}
\toprule
\multirow{2}{*}{Dataset}    & \multirow{2}{*}{Model} & \multicolumn{3}{c|}{Full Rank} & \multicolumn{3}{c}{Low Rank} \\ \cmidrule{3-8} 
                            &                        & Time [s]     & Loss      & F1     & Time [s]      & Loss     & F1     \\ \midrule
\multirow{2}{*}{Basketball} & Fixed                  & 11462 \tiny{$\pm 565$} & 0.608 \tiny{$\pm 0.00941$} & 0.685 \tiny{$\pm 0.00544$} & 2205 \tiny{$\pm 841$} & 0.849 \tiny{$\pm 0.0230$} & 0.494 \tiny{$\pm 0.00417$} \\ 
                            & \algMMT{}              & 1230 \tiny{$\pm 74.1$} & 0.699 \tiny{$\pm 0.00237$} & 0.607 \tiny{$\pm 0.00182$} & 2009 \tiny{$\pm 715$} & 0.868 \tiny{$\pm 0.0399$} & 0.475 \tiny{$\pm 0.0121$}\\ \hline
\multirow{2}{*}{Climate}    & Fixed                  & 12.5\tiny{$\pm 0.0112$} & 0.0882 \tiny{$\pm 0.0844$} & \text{-} & 269 \tiny{$\pm 319$} & 0.0803 \tiny{$\pm 0.0861$} & \text{-}  \\ 
                            & \algMMT{}              &    1.11  \tiny{$\pm 0.180$}     &  0.0825  \tiny{$\pm 0.0856$}        &     \text{-}   &   67.1 \tiny{$\pm 31.8$}        &   0.0409  \tiny{$\pm 0.00399$}    &    \text{-}    \\ 
\bottomrule
\end{tabular}
\label{tab:results}
\end{table*}

\begin{figure*}[t]
    \centering
    \begin{subfigure}[t]{0.40\linewidth}
        \includegraphics[width=\textwidth]{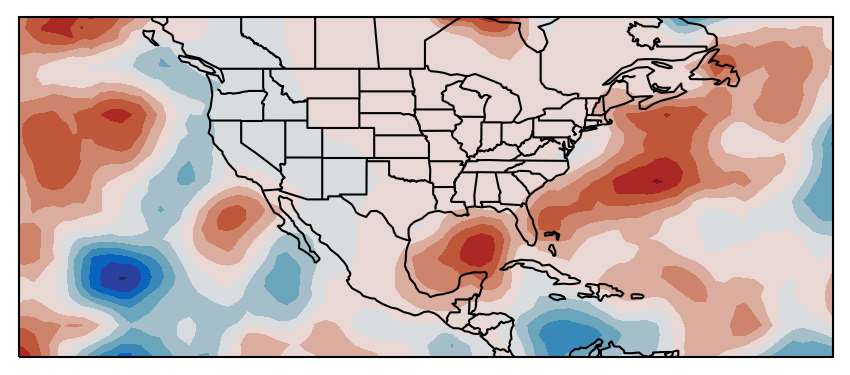}
    \end{subfigure}
    \begin{subfigure}[t]{0.40\linewidth}
        \includegraphics[width=\textwidth]{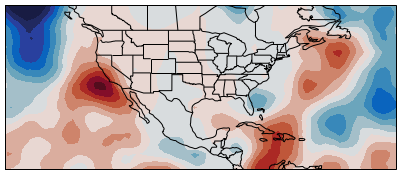}
\end{subfigure}
\caption{Climate: Some latent factors of sea surface locations after training. The red areas in the northwest Atlantic region (east of North America and Gulf of Mexico) represent areas where moisture export contributes to precipitation in the U.S. Midwest.}
\label{fig:climate_lfs}
\end{figure*}

Fig. \ref{fig:bball_multi_fixed} shows the F1 scores of \algMMT{} vs a fixed resolution model for the basketball dataset (validation loss was used as the finegraining criterion for both models). For the \full case, \algMMT{} converges 9 times faster than the fixed resolution case (the scaling of the axes obscures convergence; nevertheless, both algorithms have converged). The fixed-resolution model is able to reach a higher F1 score for the \full case, as it uses a higher resolution than \algMMT{} and is able to use more finegrained information, translating to a higher quality solution. This advantage does not transfer to the \low model.

For the \low model, the training times are comparable and both reach a similar F1 score. There is decrease in the F1 score going from \full to \low for both \algMMT{} and the fixed resolution model due to approximation error from CP decomposition. Note that this is dependent on the choice of $K$, specific to each dataset. Furthermore, we see a smaller increase in performance for the \low model vs. the \full case, indicating that the information gain from finegraining does not scale linearly with the resolution. We see a similar trend for the climate data, where \algMMT{} converges faster than the fixed-resolution model. Overall, \algMMT{} is approximately 4 $\sim$ 5 times faster and we get a similar speedup in the climate data.

\subsection{Finegraining Criteria}

We compare the performance of different finegraining criteria in Fig. \ref{fig:bball_stop_cond}. Validation loss converges much faster than other criteria for the \full model while the other finegraining criteria converge slightly faster for the \low model. In the classification case, we observe that the \full model spends many epochs training when we use gradient-based criteria, suggesting that they can be too strict for the \full case. For the regression case, we see all criteria perform similarly for the \full model, and validation loss converges faster for the \low model. As there are differences between finegraining criteria for different datasets, one should try all of them for fastest convergence.

\subsection{Interpretability}
We now demonstrate that \algMMT{} can learn semantic representations along spatial dimensions. For all latent factor figures, the factors have been normalized to $(-1,1)$ so that reds are positive and blues are negative.

\begin{figure}[ht]
\centering
    \begin{subfigure}[c]{0.22\columnwidth}
        \includegraphics[width=\textwidth]{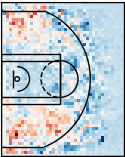}
    \end{subfigure}
    \begin{subfigure}[c]{0.23\columnwidth}
        \includegraphics[width=\textwidth]{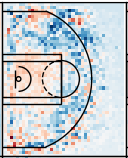}
    \end{subfigure}
    \begin{subfigure}[c]{0.22\columnwidth}
        \includegraphics[width=\textwidth]{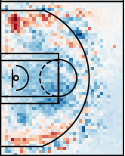}
    \end{subfigure}
\caption{
Basketball: Latent factor heatmaps of ballhandler position after training for $k=1,3,20$. They represent common shooting locations such as the right/left sides of the court, the paint, or near the three point line.}
\label{fig:bball_lfs_B_some}
\end{figure}

\begin{figure}[ht]
\centering
    \begin{subfigure}[c]{0.23\columnwidth}
        \includegraphics[width=\textwidth]{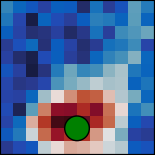}
    \end{subfigure}
    \begin{subfigure}[c]{0.22\columnwidth}
        \includegraphics[width=\textwidth]{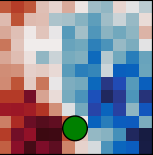}
    \end{subfigure}
    \begin{subfigure}[c]{0.22\columnwidth}
        \includegraphics[width=\textwidth]{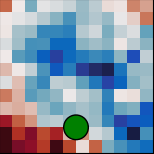}
    \end{subfigure}
\caption{
Basketball: Latent factor heatmaps of relative defender positions after training for $k=1,3,20$. The green dot represents the ballhandler at $(6, 2)$. The latent factors show spatial patterns near the ballhandler, suggesting important positions to suppress shot probability.}
\label{fig:bball_lfs_C_some}
\end{figure}

Figs. \ref{fig:bball_lfs_B_some}, \ref{fig:bball_lfs_C_some} visualize some latent factors for ballhandler position and relative defender positions, respectively (see Appendix for all latent factors). For the ballhandler position in Fig. \ref{fig:bball_lfs_B_some}, coherent spatial patterns (can be both red or blue regions as they are simply inverses of each other) can correspond to common shooting locations. These latent factors can represent known locations such as the paint or near the three-point line on both sides of the court.

For relative defender positions in Fig. \ref{fig:bball_lfs_C_some}, we see many concentrated spatial regions near the ballhandler, indicating that such close positions suppress shot probability (as expected). Some latent factors exhibit directionality as well, suggesting that guarding one side of the ballhandler may suppress shot probability more than the other side.

Fig. \ref{fig:climate_lfs} depicts two latent factors of sea surface locations. We would expect latent factors to correspond to regions of the ocean which independently influence precipitation. The left latent factor highlights the Gulf of Mexico and northwest Atlantic ocean as influential for rainfall in the Midwest due to moisture export from these regions. This is consistent with findings from \cite{li2018role, li_2016_midwest}.

\begin{figure}[t]
\centering
    \begin{subfigure}[t]{0.21\columnwidth}
        \includegraphics[width=\textwidth]{images/bball/rand_B_heatmap1.png}
    \end{subfigure}
    \begin{subfigure}[t]{0.21\columnwidth}
        \includegraphics[width=\textwidth]{images/bball/rand_B_heatmap2.png}
    \end{subfigure}
    \begin{subfigure}[t]{0.265\columnwidth}
        \includegraphics[width=\textwidth]{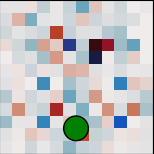}
    \end{subfigure}
    \begin{subfigure}[t]{0.265\columnwidth}
        \includegraphics[width=\textwidth]{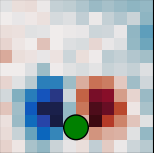}
    \end{subfigure}
\caption{
Latent factor comparisons ($k=3,10$) of randomly initialized low-rank model (1st and 3rd) and \algMMT{} (2nd and 4th) for ballhandler position (left two plots) and the defender positions (right two plots). Random initialization leads to uninterpretable latent factors.}
\label{fig:bball_lfs_rand}
\end{figure}

\paragraph{Random initialization}
\label{sub:random_init}
We also perform experiments using a randomly initialized low-rank model (without the full-rank model) in order to verify the importance of full rank initialization. Fig. \ref{fig:bball_lfs_rand} compares random initialization vs. \algMMT{} for the ballhandler position (left two plots) and the defender positions (right two plots). We observe that even with spatial regularization, randomly initialized latent factor models can produce noisy, uninterpretable factors and thus full-rank initialization is essential for interpretability.

\section{Conclusion and Future Work}
We presented a novel algorithm for tensor models for spatial analysis. Our algorithm \algMMT{} utilizes multiple resolutions to significantly decrease training time and incorporates a full-rank initialization strategy that promotes spatially coherent and interpretable latent factors. \algMMT{} is generalized to both the classification and regression cases. We proved the theoretical convergence of our algorithm for stochastic gradient descent and compared the computational complexity of \algMMT{} to a single, fixed-resolution model. The experimental results on two real-world datasets support its improvements in computational efficiency and interpretability. 

Future work includes 1) developing other stopping criteria in order to enhance the computational speedup, 2) applying our algorithm to more higher-dimensional spatiotemporal data, and 3) studying the effect of varying batch sizes between resolutions as in \citep{wu2019multigrid}.

\section*{Acknowledgements} 
This work was supported in part by NSF grants \#1850349, \#1564330, and \#1663704, Google Faculty Research Award and Adobe Data Science Research Award. We thank Stats Perform SportVU \footnote{\url{https://www.statsperform.com/team-performance/basketball/optical-tracking/}} for the basketball tracking data. We gratefully acknowledge use of the following datasets: PRISM by the PRISM Climate Group, Oregon State University \footnote{\url{http://prism.oregonstate.edu}} and EN4 by the Met Office Hadley Centre \footnote{\url{https://www.metoffice.gov.uk/hadobs/en4/}}, and thank Caroline Ummenhofer of Woods Hole Oceanographic Institution for helping us obtain the data.

\bibliography{references}
\bibliographystyle{icml2020}

\appendix
\onecolumn
\section{Theoretical Analysis}
\label{supp:theory}
\subsection{Convergence Analysis}
\label{supp:theory-conv}
\begin{theorem}\cite{bottou2018optimization}
If the step size $\eta_t \equiv \eta \leq \frac{1}{Lc_g}$, then a fixed resolution solution satisfies
\[\E[\| \V{w}_{t+1} - \V{w}_{\star}\|^2_2] \leq \gamma^t[\E[\|\V{w}_{0} - \V{w}_{\star}\|^2_2] - \beta] + \beta\]
where $\gamma = 1 - 2 \eta \mu$, $\beta=\frac{\eta \sigma_g^2}{2\mu}$, and $\V{w}_\star$ is the optimal solution.
\end{theorem}

\proof
For a single step update,
\begin{align}
    \|\V{w}_{t+1} - \V{w}_\star \|_2^2  & =  
\|\V{w}_t -\eta_t \V{g}(\V{w}_t;\xi_t) - \V{w}_\star \|_2^2  \nonumber \\
& = \|\V{w}_t - \V{w}_\star  \|_2^2 + \|\eta_t \V{g}(\V{w}_t;\xi_t) \|_2^2 - 2\eta_t \V{g}(\V{w}_t;\xi_t)(\V{w}_t-\V{w}_\star )
\end{align}
by the law of total expectation 
\begin{align}
    \E[\V{g}(\V{x}_t;\xi_t)(\V{w}_t-\V{w}_\star)] &= \E[\E[\V{g}(\V{w}_t;\xi_t)(\V{w}_t-\V{w}_\star) | \xi_{<t} ]] \nonumber \\
    &= \E[(\V{w}_t-\V{w}_\star) \E[\V{g}(\V{w}_t;\xi_t)| \xi_{<t}]] \nonumber \\
    &= \E[(\V{w}_t-\V{w}_\star)^\top \triangledown f(\V{w}_t)]
\end{align}
From strong convexity,
\begin{align}
    \langle \triangledown f(\V{w}_t) - \triangledown f(\V{w}_\star) , \V{w}_t- \V{w}_\star  \rangle  =    \langle \triangledown f(\V{w}_t) , \V{w}_t- \V{w}_\star  \rangle &\geq \mu \|\V{w}_t- \V{w}_\star\|_2^2
\end{align}
which implies $\E[(\V{w}_t-\V{w}_\star)^\top  \triangledown f(\V{w}_t)] \geq \mu\E[\|\V{w}_t- \V{w}_\star\|_2^2]$ as $\triangledown f(\V{w}_\star)=0$. Putting it all together yields
\begin{align}
     \E[\|\V{w}^{t+1} - \V{w}_\star \|_2^2 ] \leq (1-2\eta_t \mu)\E[\|\V{w}_t- \V{w}_\star\|_2^2] + (\eta_t \sigma_g)^2
     \label{eqn:single_step}
\end{align}
As $\eta_t = \eta$, we complete the contraction, by setting $\beta = \frac{(\eta \sigma_g)^2}{(2\eta \mu)}$
\begin{align}
       \E[\|\V{w}_{t+1} - \V{w}_\star \|_2^2 ] -\beta \leq (1-2\eta_t \mu) (\E[\|\V{w}_t- \V{w}_\star\|_2^2] -\beta) 
\end{align}
Repeat the iterations
\begin{align}
           \E[\|\V{w}_{t+1} - \V{w}_\star \|_2^2 ] -\beta  \leq (1-2\eta \mu)^t  (\E[\|\V{w}_{0} - \V{w}_\star\|_2^2 ] -\beta ) 
\end{align}
Rearranging the terms, we get
\begin{equation}
    \E[\|\V{w}_{t+1} - \V{w}_\star \|_2^2]  \leq (1-2\eta \mu)^t\E[\|\V{w}_{0} - \V{w}_\star ] - ((1-2\eta \mu)^t + 1)\frac{(\eta \sigma_g)^2}{(2\eta \mu)}
\end{equation}
\qed
\begin{theorem} If the step size $\eta_t \equiv \eta \leq \frac{1}{Lc_g}$, then \algMMT{} solution satisfies
\begin{align*}
    &\E[\|\V{w}^{(r)}_{t} - \V{w}^{\star}\|^2_2] \leq \gamma^{t} (\|P\|^2_{op})^r [\E[\|\V{w}^{(1)}_{0} - \V{w}^{(1),\star}\|^2_2] \\
    & \qquad - \gamma^{t} \|P\|^2_2 \beta + \gamma^{t_2} (\|P\|^2_{op}\beta - \beta) + O(1)
\end{align*}
where $\gamma = 1 - 2 \eta \mu$, $\beta=\frac{\eta \sigma_g^2}{2\mu}$, and $\|P\|_{op}$ is the operator norm of the interpolation operator $P$.
\end{theorem}
Consider a two resolution case where $R=2$ and $\V{w}^{(2)}_\star= \V{w}_\star$. Let $t_r$ be the total number of iterations of resolution $r$. Based on Eqn. \eqref{eqn:single_step}, for a fixed resolution algorithm, after $t_1+t_2$ number of iterations,
\[\E[\|\V{w}_{t_1+t_2} - \V{w}_{\star}\|_2^2 ]  -\beta \leq (1-2\eta \mu)^{t_1+t_2} (\E[\|\V{w}_{0} - \V{w}^\star \|_2^2] -\beta ) \]
For multiresolution, where we train on resolution $r=1$ first,  we have 
\begin{equation*}
E[\|\V{w}^{(1)}_{t_1} - \V{w}^{(1)}_\star \|_2^2]  - \beta \leq (1-2\eta \mu)^{t_1} (\E[\|\V{w}^{(1)}_{0} - \V{w}^{(1)}_\star\|_2^2 ] - \beta)
\end{equation*}
At resolution $r=2$, we have 
\begin{equation}
\E[\|\V{w}^{(2)}_{t_2} - \V{w}_\star \|_2^2 ]  -\beta \leq (1-2\eta \mu)^{t_2}( \E[\|\V{w}_{0}^{(2)} - \V{w}_\star \|_2^2] - \beta )
\label{eqn:multi_coarse}
\end{equation}
Using interpolation, we have $\V{w}_{0}^{(2)}=P \V{w}_{t_1}^{(1)}$. Given the spatial autocorrelation assumption, we have \[\| \V{w}_\star^{(2)} -  P \V{w}_\star^{(1)} \|_{2} \leq \epsilon\]
By the definition of  operator norm and triangle inequality,
\[\E[\|\V{w}_{0}^{(2)} - \V{w}^{(2)}_\star \|_2^2  \leq  \E [ \|P \V{w}_{t_1}^{(1)}   -   \V{w}^{(2)}_\star \|^2_2] \leq \|P\|_{op}^2 \E[ \| \V{w}_{t_1}^{(1)}   -   \V{w}^{(1)}_\star  \|_2^2  ] + \epsilon^2 \]  
Combined with eq. \eqref{eqn:multi_coarse}, we have 
\begin{align}
\E[\|\V{w}^{(2)}_{t_2} - \V{w}_\star \|_2^2 ] - \beta \leq & (1-2\eta \mu)^{t_2} (    \|P\|_{op}^2 \E[ \| \V{w}_{t_1}^{(1)}   -   \V{w}^{(1)}_\star  \|_2^2  ] + \epsilon^2 - \beta)  \\ 
 = &  (1-2\eta \mu)^{t_1+t_2} \|P\|_{op}^2 (\E[\|\V{w}^{(1)}_{0} - \V{w}^{(1)}_\star\|_2^2 ]-\beta) + (1-2\eta \mu)^{t_2} ( \|P\|_{op}^2\beta + \epsilon^2  - \beta)
 \label{eqn:two_resolution}
\end{align} 
If we initialize $\V{w}_0$ and $\V{w}^{(1)}_0$ such that $\|\V{w}^{(1)}_0-\V{w}^{(1)}_\star\|_2^2 = \|\V{w}_0-\V{w}_\star\|_2^2$,   we have \algMMT{} solution
\begin{align}
  \E[\|\V{w}^{'}_{t_1+t_2} - \V{w}_\star \|_2^2 ] - \alpha \leq   (1-2\eta \mu)^{t_1+t_2} \|P\|_{op}^2 (\E[\|\V{w}^{'}_{0} - \V{w}_\star\|_2^2 ] -\alpha)
\end{align}
for some  $\alpha$ that completes the contraction. Repeat the resolution iterates in Eqn. \eqref{eqn:two_resolution}, we reach our conclusion.
\qed

\subsection{Computational Complexity Analysis}
\label{supp:comp-proof}
In this section, we analyze the computational complexity for \algMMT{} (Algorithm \ref{alg:mrtl}). Assuming that $\nabla f$ is Lipschitz continuous, we can view gradient-based optimization as a fixed-point iteration operator $F$ with a contraction constant of $\gamma \in (0,1)$ (note that \emph{stochastic} gradient descent converges to a noise ball instead of a fixed point).
\begin{eqnarray*}
	\V{w} \leftarrow F(\V{w}), \hspace{10pt} F:=I-\eta \nabla f,
	\| F(\V{w} ) - F(\V{w}') \|\leq \gamma \| \V{w} - \V{w}'\|.
\end{eqnarray*}
Let $\V{w}^{(r)}_\star $ be the optimal estimator at resolution $r$. Suppose for each resolution $r$, we use the following finegrain criterion:
\eq{
	\| \V{w}_{t}^{(r)} - \V{w}_{t-1}^{(r)}\|  \leq \fr{C_0 D_r}{\gamma (1-\gamma)}.
	\label{supp:eqn:finegrain}
}
where $t_r$ is the number of iterations taken at level $r$. The algorithm terminates when the estimation error reaches $\fr{C_0 R}{(1-\gamma)^2}$.
The following main theorem characterizes the speed-up gained by multiresolution learning  \algMMT{}  w.r.t. the contraction factor $\gamma$ and the terminal estimation error $\epsilon$.
\begin{theorem}
	Suppose the fixed point iteration operator (gradient descent) for the optimization algorithm has a contraction factor (Lipschitz constant) of $\gamma$, the multiresolution learning procedure is faster than that of the fixed resolution algorithm by a factor of  $ \log\fr{1}{(1-\gamma) \epsilon}$, with $\epsilon$ as the terminal estimation error.
	\label{supp:thm:mmt}
\end{theorem}
We prove several useful Lemmas before proving the main Theorem \ref{supp:thm:mmt}. 
The following lemma analyzes the computational cost of the \emph{fixed-resolution} algorithm.
%
\begin{lemma}
	Given a fixed point iteration operator with a contraction factor $\gamma$, the computational complexity of a fixed-resolution training for a $p$-order tensor with rank $K$ is
	\eq{
		\mathcal{C} = \calO\brck{
			\fr{1}{|\log \gamma|} \cdot \log \left(\fr{1 }{(1-\gamma)\epsilon}\right)
			\cdot \fr{Kp}{(1-\gamma)^2\epsilon}}.
	} \label{supp:lemma:fixed}
\end{lemma}
\proof At a high level, we can prove this by choosing a small enough resolution $r$ such that the approximation error is bounded with a fixed number of iterations. Let $\V{w}_\star^{(r)}$ be the optimal estimate at resolution $r$ and $\V{w}_t$ be the estimate at step $t$. Then
\begin{equation}
\| \V{w}_\star -\V{w}_t \| \leq \| \V{w}_\star - \V{w}^{(r)}_\star \|  + \|\V{w}^{(r)}_\star - \V{w}_t  \| \leq \epsilon.
\end{equation}
We pick a fixed resolution $r$ small enough such that
\eq{\| \V{w}_\star-\V{w}_\star^{(r)}\| \leq \fr{\epsilon}{2},}
then using the termination criteria $\| \V{w}_\star -\V{w}^{(r)}_\star \| \leq \fr{C_0 R}{(1-\gamma)^2}$ gives $D_r = \Omega ((1-\gamma)^2\epsilon)$ where $D_r$ is the discretization size at resolution $r$. Initialize $\V{w}_0=0$ and apply $F$ to $\V{w}$ for $t$ times such that
\eq{
	\fr{\gamma^t}{2(1-\gamma) } \|F(\V{w}_0) \| \leq \fr{\epsilon}{2}.
}
As $\V{w}_0=0$, $\|F(\V{w}_0) \| \leq 2C$, we obtain that
\eq{
	t\leq  \fr{1}{|\log \gamma|} \cdot \log \brck{\fr{2C}{(1-\gamma)\epsilon}},
}
Note that for an order $p$ tensor with rank $K$, the computational complexity of every iteration in \algMMT{} is $ \calO(Kp/D_r)$ with $D_r$ as the discretization size. Hence, the computational complexity of the fixed resolution training is
\eq{
	\mathcal{C}
	&= \calO\brck{
		\fr{1}{|\log \gamma|} \cdot \log \brck{\fr{1}{ (1-\gamma)\epsilon}}
		\cdot \fr{Kp}{D_r}
	} \notag\\
	&= \calO\brck{
		\fr{1}{|\log \gamma|} \cdot \log \brck{\fr{1 }{(1-\gamma)\epsilon}}
		\cdot \fr{Kp}{(1-\gamma)^2\epsilon}
	}. \notag\qed
}

Given a spatial discretization $r$, we can construct an operator $F_r$ that learns discretized tensor weights. The next lemma relates the estimation error with resolution. The following lemma relates the estimation error with resolution:

\begin{lemma}\citep{nash2000multigrid}
	For each resolution level $r = 1,\cdots,R$, there exists a constant $C_1$ and $C_2$,	such that the fixed point iteration with discretization size $D_r$ has an estimation error:
	\eq{
		\|F(\V{w}) - F^{(r)}(\V{w})\|\leq (C_1 + \gamma C_2 \|\V{w}\|) D_r}
	\label{supp:lemma:disc}
\end{lemma}
\proof 
See \citep{nash2000multigrid} for details.

We have obtained the discretization error for the fixed point operation at any resolution. Next we analyze the number of iterations $t_r$ needed at each resolution $r$ before finegraining.
\begin{lemma}
    For every resolution $r=1,\dots,R$, there exists a constant $C'$ such that the number of iterations $t_r$ before finegraining satisfies:
	\eq{
		t_r \leq C' / \log | \gamma |
	}
	\label{supp:lemma:iter_level}
\end{lemma}
\proof%
According to the fixed point iteration definition, we have for each resolution $r$:
\begin{eqnarray}
	\|F_r(\V{w}_{t_r}) - \V{w}_{t_r}^{(r)}) \| &\leq& \gamma^{t_r-1} \| F_r(\V{w}^{(r)}_0) - \V{w}^{(r)}_0\| \\ 
	&\leq& \gamma^{t_r-1} \frac{C_0 D_r}{1-\gamma}\\
	&\leq &  C'\gamma^{t_r-1} 
\end{eqnarray}
using the definition of the finegrain criterion. \qed

By combining Lemmas \ref{supp:lemma:iter_level} and the computational cost per iteration, we can compute the total computational cost  for our \algMMT{} algorithm, which is proportional to the total number of iterations for all resolutions:
\eq{
	\mathcal{C}_{\algMMT{}} &= \calO\brck{\fr{1}{|\log \gamma|}\brcksq {(D_r/Kp)^{-1} +(2 D_r/Kp)^{-1} + (4 D_r/Kp)^{-1} + \cdots} } \notag\\
	&=\calO\brck{\fr{1 }{|\log \gamma| }\brck{\fr{Kp}{D_r}}\brcksq{1 + \fr{1}{2} + \fr{1}{4} +\cdots} } \notag\\
	&=\calO\brck{\fr{1 }{|\log \gamma|} \brck{ \fr{Kp}{D_r} } \brcksq{ \fr{1-(\fr{1}{2}) ^{n}}{1-\fr{1}{2}} } } \notag\\
	&= \calO\brck{\fr{1 }{|\log \gamma|} \brck{\fr{Kp}{(1-\gamma)^2\epsilon}}},
}
where the last step uses the termination criterion in (\ref{supp:eqn:finegrain}). Comparing with the complexity analysis for the fixed resolution algorithm in Lemma \ref{supp:lemma:fixed}, we complete the proof.
\qed

\section{Experiment Details}
\label{supp:experiments}

\paragraph{Basketball}
We list implementation details for the basketball dataset. We focus only on half-court possessions, where all players have crossed into the half court as in \cite{Yue2014}. The ball must also be inside the court and be within a 4 foot radius of the ballhandler. We discard any passing/turnover events and do not consider frames with free throws.

For the ball handler location $\{D_r^1\}$, we discretize the half-court into resolutions $4 \times 5, 8 \times 10, 20 \times 25, 40 \times 50$. For the relative defender locations, at the full resolution, we choose a $12 \times 12$ grid around the ball handler where the ball handler is located at $(6, 2)$ (more space in front of the ball handler than behind him/her). We also consider a smaller grid around the ball handler for the defender locations, assuming that defenders that are far away from the ball handler do not influence shooting probability. We use $6 \times 6 ,12 \times 12$ for defender positions.

Let us denote the pair of resolutions as $(D_r^1, D_r^2)$. We train the full-rank model at resolutions $(4 \times 5, 6 \times 6), (8 \times 10, 6 \times 6), (8 \times 10, 12 \times 12)$ and the low-rank model at resolutions $(8 \times 10, 12 \times 12), (20 \times 25, 12 \times 12), (40 \times 50, 12 \times 12)$.

There is a notable class imbalance in labels (88\% of data points have zero labels) so we use weighted cross entropy loss using the inverse of class counts as weights. For the low-rank model, we use tensor rank $K=20$. The performance trend of \algMMT{} is similar across a variety of tensor ranks. $K$ should be chosen appropriately to the desired level of approximation.

\paragraph{Climate}
We describe the data sources used for climate. The precipitation data comes from the PRISM group \cite{prism}, which provides estimates monthly estimates at 1/24º spatial resolution across the continental U.S from 1895 to 2018. For oceanic data we use the EN4 reanalysis product \cite{en4}, which provides monthly estimates for ocean salinity and temperature at 1º spatial resolution across the globe from 1900 to the present (see Fig. \ref{fig:climate_data}). We constrain our spatial analysis to the range [-180ºW, 0ºW] and [-20ºS, 60ºN], which encapsulates the area around North America and a large portion of South America.

The ocean data is non-stationary, with the variance of the data increasing over time. This is likely due to improvement in observational measurements of ocean temperature and salinity over time, which reduce the amount of interpolation needed to generate an estimate for a given month. After detrending and deseasonalizing, we split the train, validation, and test sets using random consecutive sequences so that their samples come from a similar distribution.

We train the full-rank model at resolutions $4\times 9$ and $8 \times 18$ and the low-rank model at resolutions $8\times18$, $12 \times 27$, $24 \times 54$, $40 \times 90$, $60 \times 135$, and $80 \times 180$. For finegraining criteria, we use a patience factor of 4, i.e. training was terminated when a finegraining criterion was reached a total of 4 times. Both validation loss and gradient statistics were relatively noisy during training (possibly due to a small number of samples), leading to early termination without the patience factor.

During finegraining, the weights were upsampled to the higher resolution using bilinear interpolation and then scaled by the ratio of the number of inputs for the higher resolution to the number of inputs for the lower resolution (as described in Section \ref{sec:mrtl}) to preserve the magnitude of the prediction. 

\paragraph{Details}
We trained the basketball dataset on 4 RTX 2080 Ti GPUs, while the climate dataset experiments were performed on a separate workstation with 1 RTX 2080 Ti GPU. The computation times of the fixed-resolution and \algMMT{} model were compared on the same setup for all experiments.

\subsection{Hyperparameters}
\begin{table}[h]
  \centering
  \footnotesize
\begin{tabular}{l|l|l}
Hyperparameter                     & Basketball          & Climate             \\ \hline
Batch size                         & $32 - 1024$         & $8 - 128$           \\
Full-rank learning rate $\eta$     & $10^{-3} - 10^{-1}$ & $10^{-4} - 10^{-1}$ \\
Full-rank regularization $\lambda$ & $10^{-5} - 10^{0}$  & $10^{-4} - 10^{-1}$ \\
Low-rank learning rate $\eta$      & $10^{-5} - 10^{-1}$ & $10^{-4} - 10^{-1}$ \\
Low-rank regularization $\lambda$  & $10^{-5} - 10^{0}$  & $10^{-4} - 10^{-1}$ \\
Spatial regularization $\sigma$    & $0.03 - 0.2$        & $0.03 - 0.2$        \\
Learning rate decay $\gamma$       & $0.7 - 0.95$        & $0.7 - 0.95$           
\end{tabular}
  \caption{Search range for \algSGD{} hyperparameters}
  \label{supp:tab_hyperp}
\end{table}
Table \ref{supp:tab_hyperp} show the search ranges of all hyperparameters considered. We performed separate random searches over this search space for \algMMT{}, fixed-resolution model, and the randomly initialized low-rank model. We also separate the learning rate $\eta$ and regularization coefficient $\lambda$ between the full-rank and low-rank models.

\subsection{Accuracy and Convergence}

\begin{figure}[ht]
    \centering
    \begin{subfigure}[t]{0.49\linewidth}
        \includegraphics[width=\textwidth]{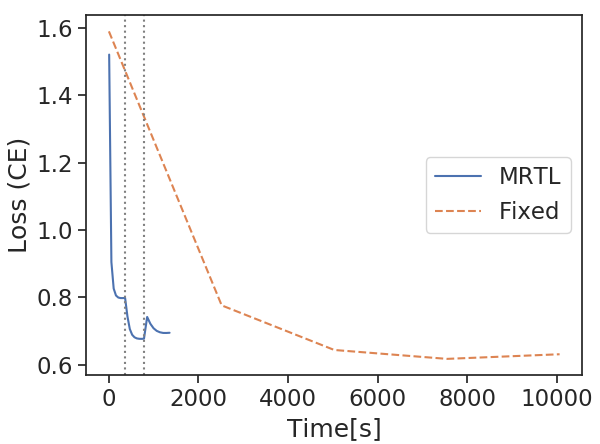}
    \end{subfigure}
    \begin{subfigure}[t]{0.49\linewidth}
        \includegraphics[width=\textwidth]{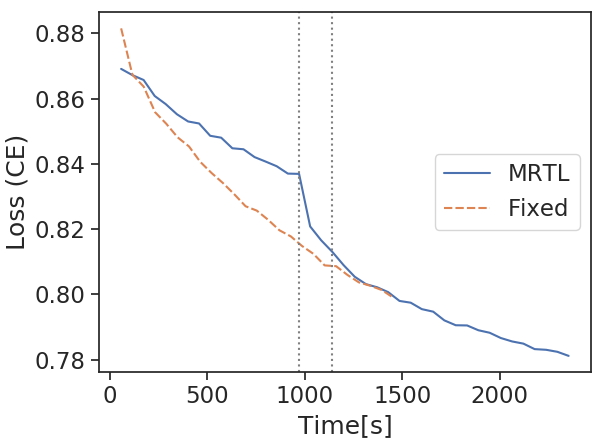}
    \end{subfigure}
\caption{Basketball: Loss curves of \algMMT{} vs. the fixed-resolution model for the \full (left) and \low model (right). The vertical lines indicate finegraining to the next resolution.}
\label{fig:bball_multi_fixed_loss}
\end{figure}
Fig. \ref{fig:bball_multi_fixed_loss} shows the loss curves of \algMMT{} vs. the fixed resolution model for the \full and \low case. They show a similar convergence trend, where the fixed-resolution model is much slower than \algMMT{}.

\subsection{Finegraining Criteria}

Table \ref{tab:stop_cond} lists the results for the different finegraining criteria. In the classification case, we see that validation loss reaches much faster convergence than other gradient-based criteria in the full-rank case, while the gradient-based criteria are faster for the low-rank model. All criteria can reach similar F1 scores. For the regression case, all stopping criteria converge to a similar loss in roughly the same amount of time for the full-rank model. For the low-rank model, validation loss appears to converge more quickly and to a lower loss value. 

\begin{table}[h]
\footnotesize
\centering
\captionsetup{justification=centering}
\caption{Runtime and prediction performance comparison of different finegraining criteria}
\begin{tabular}{l|l|c|c|c|c|c|c}
\toprule
\multirow{2}{*}{Dataset}    & \multirow{2}{*}{Model} & \multicolumn{3}{c|}{Full-Rank} & \multicolumn{3}{c}{Low-Rank} \\ \cmidrule{3-8} 
                            &                        & Time [s]     & Loss      & F1     & Time [s]      & Loss     & F1     \\ \midrule
\multirow{4}{*}{Basketball} & Validation loss              & \textbf{1230 \tiny{$\pm 74.1$}} & 0.699 \tiny{$\pm 0.00237$} & 0.607 \tiny{$\pm 0.00182$} & 2009 \tiny{$\pm 715$} & 0.868 \tiny{$\pm 0.0399$} & 0.475 \tiny{$\pm 0.0121$}\\ 
                            & Gradient norm         & 7029 \tiny{$\pm 759$} & 0.703 \tiny{$\pm 0.00216$} & \textbf{0.610 \tiny{$\pm 0.00149$}}                & \textbf{912 \tiny{$\pm 281$}}  & 0.883 \tiny{$\pm 0.00664$}       & 0.476 \tiny{$\pm 0.00270$}     \\
                            & Gradient variance                           & 7918 \tiny{$\pm 1949$} & 0.701 \tiny{$\pm 0.00333$} & 0.609 \tiny{$\pm 0.00315$}                & 933 \tiny{$\pm 240$}  & 0.883 \tiny{$\pm 0.00493$}       & \textbf{0.476 \tiny{$\pm 0.00197$}}     \\
                            & Gradient entropy                            & 8715 \tiny{$\pm 957$}  & 0.697 \tiny{$\pm 0.00551$} & 0.597 \tiny{$\pm 0.00737$}                & 939 \tiny{$\pm 259$}  & 0.886 \tiny{$\pm 0.00248$}       & 0.475 \tiny{$\pm 0.00182$}     \\ \hline
\multirow{4}{*}{Climate}    & Validation loss              &    1.04  \tiny{$\pm 0.115$}     &  \textbf{0.0448}  \tiny{$\pm 0.0108$}        &     \text{-}   &   \textbf{37.4} \tiny{$\pm 28.7$}        &   \textbf{0.0284}  \tiny{$\pm 0.00171$}    &    \text{-}    \\ 
                            & Gradient norm         &  1.11 \tiny{$\pm 0.0413$}  &     0.0506 \tiny{$\pm 0.00853$}     &     \text{-}   &    59.1 \tiny{$\pm 16.9$}   &   0.0301 \tiny{$\pm 0.00131$}   &    \text{-}    \\ 
                            & Gradient variance              &   1.14 \tiny{$\pm 0.0596$}   &  0.0458 \tiny{$\pm 0.00597$}   &     \text{-}   &  62.9 \tiny{$\pm 14.4$}   &  0.0305 \tiny{$\pm 0.00283$}  &    \text{-}    \\ 
                            & Gradient entropy              & \textbf{0.984 \tiny{$\pm 0.0848$}}  &  0.0490 \tiny{$\pm 0.0144$}  &     \text{-}   &     48.4 \tiny{$\pm 21.1$}  & 0.0331 \tiny{$\pm 0.00949$} &    \text{-}    \\ 
                            
\bottomrule
\end{tabular}
\label{tab:stop_cond}
\end{table}

\subsection{Random initialization}

\begin{figure}[!h]
\centering
    \begin{subfigure}[c]{0.35\columnwidth}
        \includegraphics[width=\textwidth]{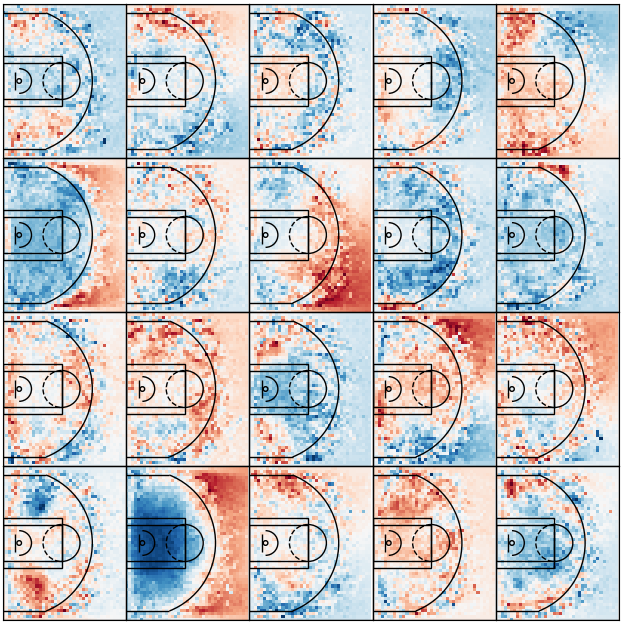}
    \end{subfigure}
    \begin{subfigure}[c]{0.35\columnwidth}
        \includegraphics[width=\textwidth]{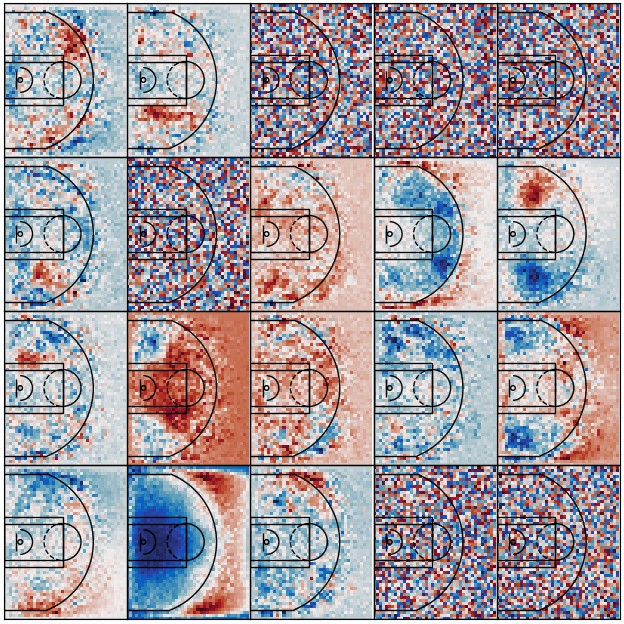}
    \end{subfigure}
\caption{
Basketball: Latent factors of ball handler position after training \algMMT{} (left) and a low-rank model using random initialization (right). The factors have been normalized to (-1,1) so that reds are positive and blues are negative. The latent factors are numbered left to right, top to bottom.}
\label{fig:sup_bball_lfs_B}
\end{figure}

\begin{figure}[!h]
\centering
    \begin{subfigure}[c]{0.35\columnwidth}
        \includegraphics[width=\textwidth]{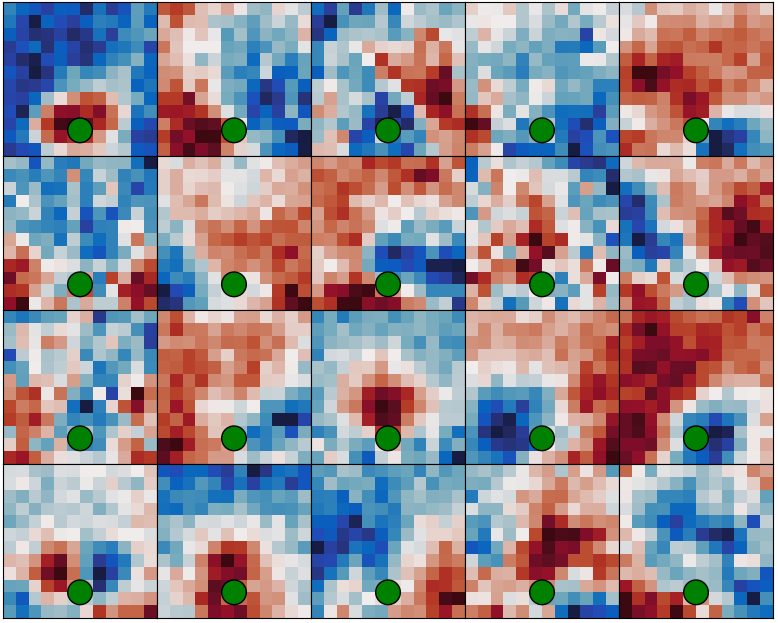}
    \end{subfigure}
    \begin{subfigure}[c]{0.35\columnwidth}
        \includegraphics[width=\textwidth]{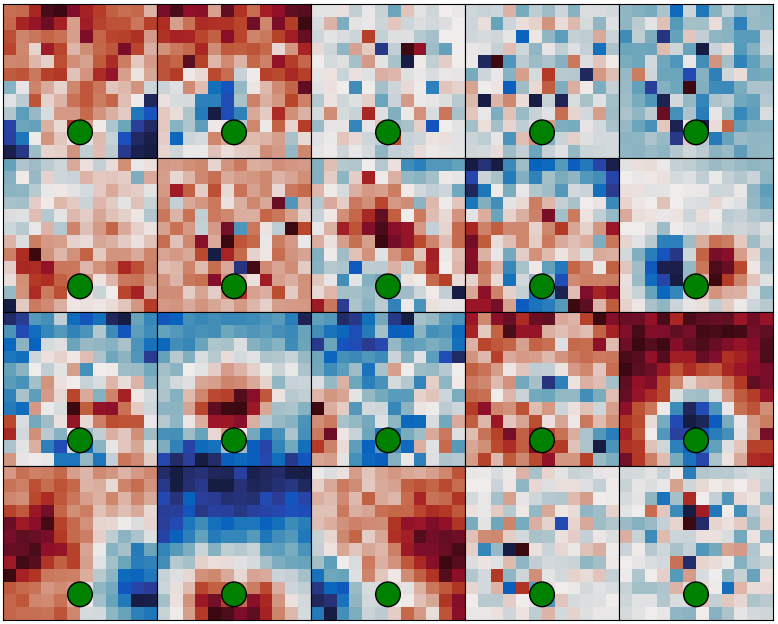}
    \end{subfigure}
\caption{
Basketball: Latent factors of relative defender positions after training \algMMT{} (left) and a low-rank model using random initialization (right). The factors have been normalized to (-1,1) so that reds are positive and blues are negative. The green dot represents the ballhandler at (6, 2). The latent factors are numbered left to right, top to bottom.}
\label{fig:sup_bball_lfs_C}
\end{figure}

Fig. \ref{fig:sup_bball_lfs_B} shows all latent factors after training \algMMT{} vs a randomly initialized low-rank model for ballhandler position. We can see clearly that full-rank initialization produces spatially coherent factors while random initialization can produce some uninterpretable factors (e.g. the latent factors for $k=3,4,5,7,19,20$ are not semantically meaningful). Fig. \ref{fig:sup_bball_lfs_C} shows latent factors for the defender position spatial mode, and we can draw similar conclusions about random initialization.

\end{document}